\documentclass[10pt,twocolumn,letterpaper]{article}

\usepackage{cvpr}              
\usepackage{url}
\usepackage[dvipsnames]{xcolor}

\newcommand{\boxedthm}[1]{
\begin{tcolorbox}[colback=gray!30,
                  colframe=black,
                  width=\linewidth,
                  arc=2mm, auto outer arc,
                  boxrule=1pt,
                  boxsep=-1mm,
                 ]
  #1
\end{tcolorbox}
\providecommand{\vY}{\mathbf{Y}}
}

\usepackage{multirow}
\usepackage{booktabs}

\newtheorem{theorem}{Theorem}[section]

\DeclareMathOperator{\E}{\mathbb{E}}
\usepackage[most]{tcolorbox}

\definecolor{cvprblue}{rgb}{0.21,0.49,0.74}
\usepackage[pagebackref,breaklinks,colorlinks,citecolor=cvprblue]{hyperref}

\def\paperID{8651} 
\def\confName{CVPR}
\def\confYear{2024}

\title{Generative Quanta Color Imaging}

\author{Vishal Purohit, Junjie Luo, Yiheng Chi, Qi Guo, Stanley H. Chan, Qiang Qiu\\
Elmore Family School of Electrical and Computer Engineering\\
Purdue University, West Lafayette, Indiana\\
{\tt\small \{purohitv,luo330,chi14,guo675,stanchan,qqiu\}@purdue.edu}
}

\begin{document}
\maketitle  

\begin{abstract}
The astonishing development of single-photon cameras has created an unprecedented opportunity for scientific and industrial imaging. However, the high data throughput generated by these 1-bit sensors creates a significant bottleneck for low-power applications. In this paper, we explore the possibility of generating a color image from a single binary frame of a single-photon camera. We evidently find this problem being particularly difficult to standard colorization approaches due to the substantial degree of exposure variation. The core innovation of our paper is an exposure synthesis model framed under a neural ordinary differential equation (Neural ODE) that allows us to generate a continuum of exposures from a single observation. This innovation ensures consistent exposure in binary images that colorizers take on, resulting in notably enhanced colorization. We demonstrate applications of the method in single-image and burst colorization and show superior generative performance over baselines. Project website can be found at this \href{https://vishal-s-p.github.io/projects/2023/generative_quanta_color.html}{link.}
\end{abstract} 
\vspace{-4ex}

\section{Introduction}
\label{sec:intro}

Single-photon image sensors are gaining a strong momentum over the past decade due to their impeccable photon-counting capability. This not only opens the door for advanced scientific imaging in low-light environment, but also creates the opportunity for new imaging in high-speed, high dynamic range, and low-bit conditions. Single-photon imaging today are realized through single-photon avalanche diodes (SPAD) and quanta image sensors (QIS). The latest development of QIS has demonstrated a mega-pixel resolution~\cite{ma2017photon} which is compatible with mainstream cell phone cameras, whereas SPAD has achieved hundreds of thousands of (binary) frames per second which has enabled a plethora of industrial and 3D imaging applications~\cite{dutton16qvga}. 
\begin{figure}[t]
 \centering
  \includegraphics[width=\columnwidth]{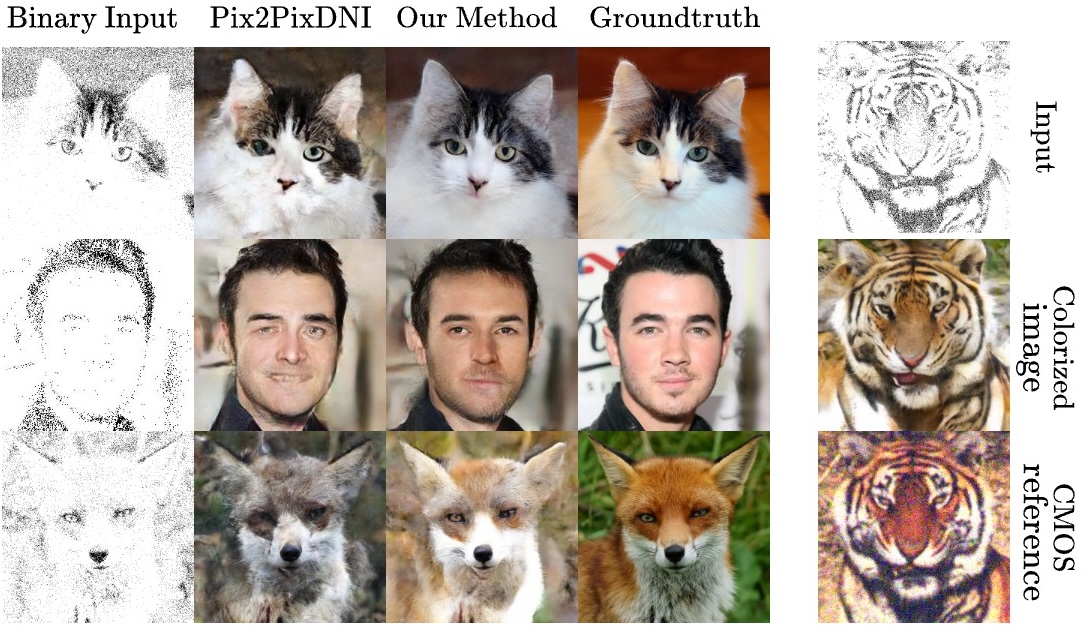}
  \caption{We introduce generative quanta color imaging. Given a binary frame captured by a single-photon camera (quanta image sensor
(QIS) in this example), the proposed method generates a continuum of exposures using a neural ordinary differential equation framework.
 Color are then generated based on these exposures. \textbf{Left:} A qualitative comparison between our approach and existing methods is shown. \textbf{Right:} Comparison between colorization results and image captured using a RGB CMOS camera under similar conditions.}
  \vspace{-10pt}
  \label{fig:fig_eg_spect}
\end{figure}

\textbf{Problem Statement}. In the context of 1-bit single-photon imaging, the data volume generated is huge. Taking a $1000\times1000$ SPAD array as an example, a video of 500 binary frames requires a 500 Mb/s data transfer rate; if the frame rate becomes 150k fps, the data transfer rate increases to 150 Gb/s. Since data throughput scales proportionally with the power consumption, it quickly becomes infeasible to use these sensors continuously without considering power supply, memory, and transmission issues. While solid progress are being made across the sensor and signal processing community~\cite{ma15,ma15high,ma15pump, ma21rms,burri14,dutton16,chi20,gnanasambandam20-review,gana2022,ma22-review}, 
the data throughput remains an unsolved problem for single-photon sensors to be considered applicable in many applications, including low-power devices, wireless devices, underwater vehicles, spacecrafts, etc., where bandwidth is the bottleneck.

In this paper, instead of compressing data or redesigning the sensor, we ask the question

\boxedthm{
\begin{center}
Is it possible to recover a \underline{color} image from a \emph{single} 1-bit image?
\end{center}
}

\textbf{Significance}. There are several reasons why this problem is significant. Firstly, while there is an ocean of literature on image generation (through methods such as Generative Adversarial Networks~\cite{goodfellow14} and diffusion~\cite{ho2020denoising}), our problem is unique in the sense none of the existing work addresses it. Besides being binary (and hence colorless), in practice the input data usually also suffers from under- or over-exposures. Therefore, before generating color, it is even more critical to synthesize the unseen exposure. Secondly, any solution offered to our problem will provide a path to generative imaging in the future. From the point of view of compressed acquisition, our method demonstrates the extreme case of compression where the incoming data is a single bit. For future AR/VR applications where power is severely limited, our approach shows the possibility of extreme data compression.

\textbf{Contributions}. The core algorithmic contribution of this paper is two-fold
\begin{itemize}
\item We propose a neural ordinary differential equation (neural ODE) based framework to synthesize a continuum of exposures that are not available in the measurement. By controlling the integration interval of neural ODE, we obtain convolutional filters that are responsible for converting input representation to image representation of desired exposure. 
\item We theoretically and empirically show that a controlled variation of filter atoms leads to controlled exposure changes in the output image.
\end{itemize}
We verify our proposed method using both synthetic and real single-photon image sensor data. A snapshot of the results is shown in Fig.~\ref{fig:fig_eg_spect}.

\section{Related Work}\label{sec:background}
\begin{figure*}
    \centering
    \includegraphics[width=\textwidth]{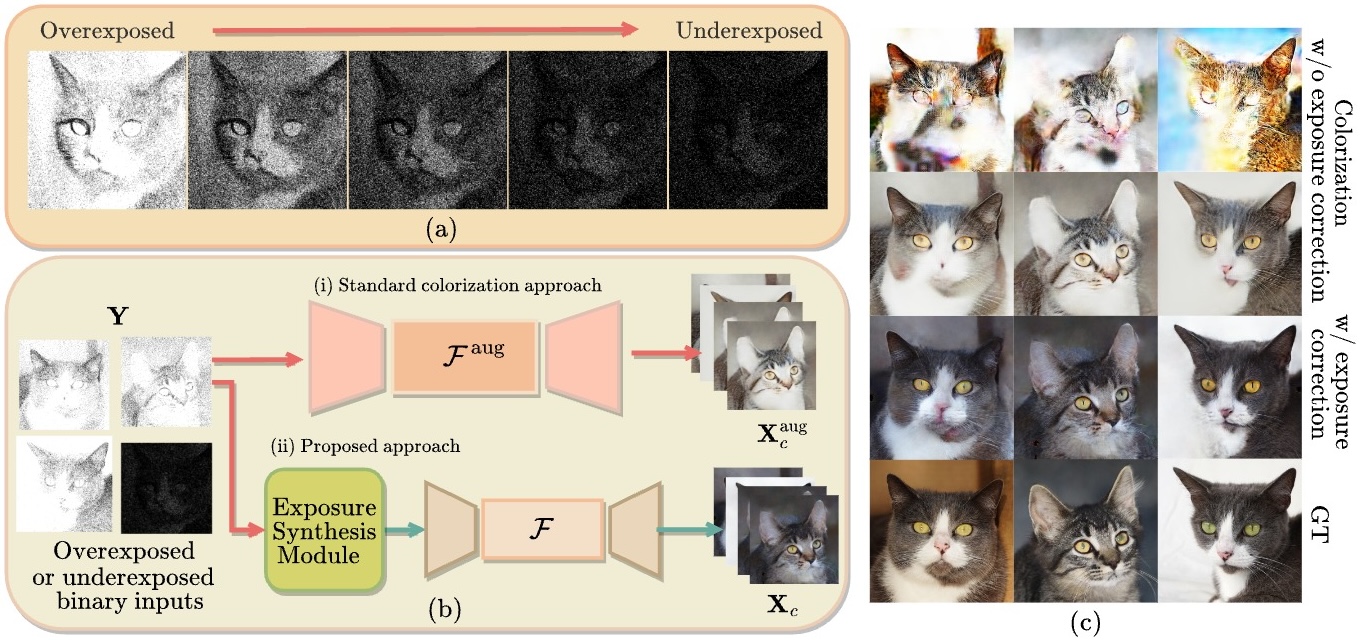}
    \caption{Illustration of exposure correction and colorization of binary images using neural networks. (a) depicts a range of images from overexposed to underexposed, illustrating the degradation of image details due to exposure variation. (b) contrasts the standard colorization workflow and our proposed approach. (\romannumeral 1) In standard colorization approaches, a neural network learns to map a binary image $\mathbf{Y}$ to the corresponding color image $\mathbf{X}_{c}$ via a neural network, $\mathcal{F}^{\text{aug}}$, where superscript `aug' indicates the colorizer is trained using dataset with augmented exposure images.  (\romannumeral 2) In contrast, our approach does not require training colorizer with augmented exposure images. (c) compares the colorization results: the first row is the output of a colorizer trained without augmentation, the second row is the output of colorizer trained with augmented data, the third row corresponds to the results of our method and the last provides the ground truth images for reference.}
    \label{fig:colorize pipelines}
\end{figure*}
\textbf{Single-Photon Imaging}. The context of this paper is single-photon image sensors, of which we are mostly interested in \emph{one-bit} sensing because of its potential in high-speed and high dynamic range imaging~\cite{chan22-1bqis}. Today, 1-bit sensing has been realized using quanta image sensors (QIS)~\cite{fossum05crc-stillcam, fossuem13, yin2022threshold, fossum2022analog, ma22-review} and single-photon avalanche diodes (SPAD)~\cite{burri14,dutton16qvga,dutton16}. From an image reconstruction point of view, since 1-bit data contains so little information, all existing image reconstruction methods need to use multiple frames. For QIS, these methods range from classical optimization~\cite{chan2014efficient, Chan16-admm}, transform-denoise~\cite{chan16-bits}, neural network~\cite{Choi18}, color demosaicking~\cite{elgendy2021low}, frequency demodulation~\cite{elgendy2021low}, and student-teacher learning~\cite{gnanasambandam20-dark, chi20}. For SPAD, quanta burst photography~\cite{Ma21} and its variants \cite{gyongy2018single, seets2021motion, iwabuchi2021image, Ingle21} first align frames and then apply image fusions~\cite{Gupta19, Gupta19flood,Barragan22,Sundar22}. The question we ask in this paper is to push the input requirement to the limit: can we recover/generate a color image from a binary input?

\noindent\textbf{Image-to-Image Translation}. This paper focuses on using generative models. Generative Adversarial Networks (GANs)~\cite{goodfellow14} have been a significant breakthrough in the field of image translation. Pix2Pix~\cite{isola2017image} and CycleGAN~\cite{zhu17} demonstrated the effectiveness of GANs in image-to-image translation tasks, with paired and unpaired datasets, respectively. Several subsequent works have focused on improving image fidelity, disentangling style and content, and generating images with varying styles, such as Pix2PixHD~\cite{wang18-cgan}, StarGAN~\cite{choi2018stargan}, StyleGAN~\cite{karras21}, and StyleGANv2~\cite{choi2020starganv2}. Recent advancements in continuous cross-domain translation techniques have further improved the output quality by leveraging images from intermediate domains. These techniques include interpolation of networks~\cite{lin19-realgan, lira20-ganhopper}, interpolation of two latent vectors~\cite{ choi2020starganv2, karras21}, and exploiting the interpolation path and translation of manifold~\cite{chen19-homomor,liu21, pizzati21-comogan}.

\noindent\textbf{Convolutional Filter Decomposition}. In the proposed method, one of the key components is the convolutional filter decomposition. Several techniques have been proposed to decompose filters of Convolutional Neural Networks (CNNs) in order to reduce the computation complexity by exploiting the redundancy in convolutional filters~\cite{cheng2018rotdcf,lebedevGROL14,miao2021-3d,osti_10330352, wangthesis, miao2022continual}. Filter decomposition methods achieve similar performance as compared to the original CNNs while offering speedups in computation. As it is expensive to model the high-dimension space of convolutional filters directly, Qiu \etal \cite{qiu18dcfnet} observed that the convolutional filters can often be approximated using a linear combination small set of basis elements. By decomposing filters into a small set of basis, Wang \etal \cite{wang20stochastic, wang2021image} were able to sample the basis and demonstrated theoretical and empirical results on stochastic image generation tasks. Futher, Chen \etal~\cite{cheninner2023} showed that filter decomposition can be used to calculate network similarity and large convolutional models can be fine tuned using filter decomposition~\cite{chen2024large}. 
\section{Proposed Approach}\label{sec:proposed_approach}
    \subsection{One-Bit Image Formation Model }\label{subsec:img_formation}
    We consider a simplified image formation model for 1-bit image sensors. Letting $\theta$ be the underlying quanta exposure \cite{fossuem13}, the observed signal is defined by
    \begin{equation*}
        Y = \text{ADC}\Big\{ \text{Poisson}(\theta) + \text{Gaussian}(0,\sigma_r^2) \Big\},
    \end{equation*}
    where $\sigma_r$ denotes the standard deviation of read noise, and ADC is the analog-to-digital conversion. For 1-bit sensing, the ADC can be modeled as a threshold such that $\text{ADC}(x) = 1$ if $x \ge q$ and $\text{ADC}(x) = 0$ if $x < q$ for some threshold $q$. Since $Y$ is a binary random variable, it is fully characterized by the probability mass function, which is \cite{chan22-1bqis}
    \begin{equation*}
        p_Y(1) = \sum_{k=0}^{\infty} \frac{e^{-\theta}\theta^k}{k!}\Phi\left(\frac{k-q}{\sigma_r}\right),
    \end{equation*}
    where $\Phi(z) = \int_{-\infty}^z \frac{1}{\sqrt{2\pi}}\exp\{-t^2/2\}dt$ is the cumulative distribution of a standard Gaussian. It then follows that the mean (or the bit density) is $\mu = \E[Y] = p_Y(1)$.

    \begin{table}[ht]
        \centering
        \resizebox{\columnwidth}{!}{%
        \begin{tabular}{@{}lcccc@{}}
            \toprule
            \textbf{Method} & \textbf{Generator Size} & \textbf{Dataset Size} & \textbf{Training Time} & \textbf{FID} \\
            \midrule
            Standard colorization & ResNet-9 & 3x & 1x & 62.34 \\
            + Increased training time & ResNet-9 & 3x & 1.5x & 54.25 \\
            + Increased model size & ResNet-10 & 3x & 1x & 74.58 \\
            Proposed method & ResNet-9 & 1x & 1x & \textbf{49.88} \\
            \bottomrule
        \end{tabular}%
        }
        \caption{Performance comparison of the standard colorization approach under various training conditions versus the proposed method.}
        \label{tab:train_conditions}
        \end{table}

    \subsection{Why Exposure Synthesis is Required for Colorization?}\label{subsec:exp_synth}
    Given a binary image $\mathbf{Y} \in \mathbb{R}^{N}$, our goal is to recover a color image $\mathbf{X}_c \in \mathbb{R}^{N \times 3}$. However, since the measurement does not contain any color information, our problem fits the classical work of image colorization~\cite{vitoria2020chromagan, Su-CVPR-2020, wu2021towards, kumar2021colorization}. However, it is non-trivial to deploy such an extension in real world scenarios, because very likely there is a discrepancy in the exposure levels of the input images compared to those used during training. 
    
    An example of binary images of different exposures are shown in Figure \ref{fig:colorize pipelines}(a) illustrating the effect of exposure change on the details in the captured binary binary image. Consider a typical colorization network for example, an Image-to-Image translation model like Pix2Pix~\cite{pix2pix2017} trained on well exposed image but tested on overexposed images, the colorization results are notably poorer (shown in Figure \ref{fig:colorize pipelines}(c) row 1). This is a realistic concern because the underlying exposure $\boldsymbol{\theta}$ can fluctuate dramatically in real life. 
     
     \noindent Possible solutions include (\romannumeral 1) \textbf{Exposure Augmentation :} training dataset is enriched with images of varying exposures, at a cost of additional training time. We find, however, this approach partially addresses the issue, as demonstrated in our results shown in Figure \ref{fig:colorize pipelines}(b) row 2. In particular, this approach falls short at over- or under-exposure.
     One can alleviate this issue by applying multiple colorizers, each handling a narrow range of exposures, but this is not ideal as it demands substantially more computational resources. (\romannumeral 2) \textbf{Increased Model size :}  Another solution is to increase the size of generator used in Pix2Pix~\cite{pix2pix2017} and train for longer period of time. In Table \ref{tab:train_conditions}, we present the colorization performance represented by the FID metric under various training scenarios. Under increased generator size or increased training time the standard colorization approach performs poorly compared to proposed method detailed in subsequent sections. 
   
     \textbf{Proposed Solution.} We address the challenge by a dedicated exposure correction based on neural ordinary differential equations (neural ODEs) to standardize the exposure of binary images.
     This specialization allows the colorizer to focus on consistent-exposure images, resulting in notably enhanced colorization, as evidenced Figure \ref{fig:colorize pipelines}(c) row 3. Additionally, our exposure synthesis can generate multiple exposures from a single binary image, enabling more effective multi-exposure colorization. Some of the added advantages of our approach include: (a) \textbf{Modularity}. The colorizer in our design is interchangeable, allowing for upgrades to more advanced architectures without requiring retraining on a large, augmented dataset. (b) \textbf{Parameter Efficiency}. Our method achieves its objectives without resorting to multiple networks with large parameter sets, ensuring a more streamlined and resource-efficient process.
    \begin{figure*}
        \centering
        \includegraphics[width=\textwidth]{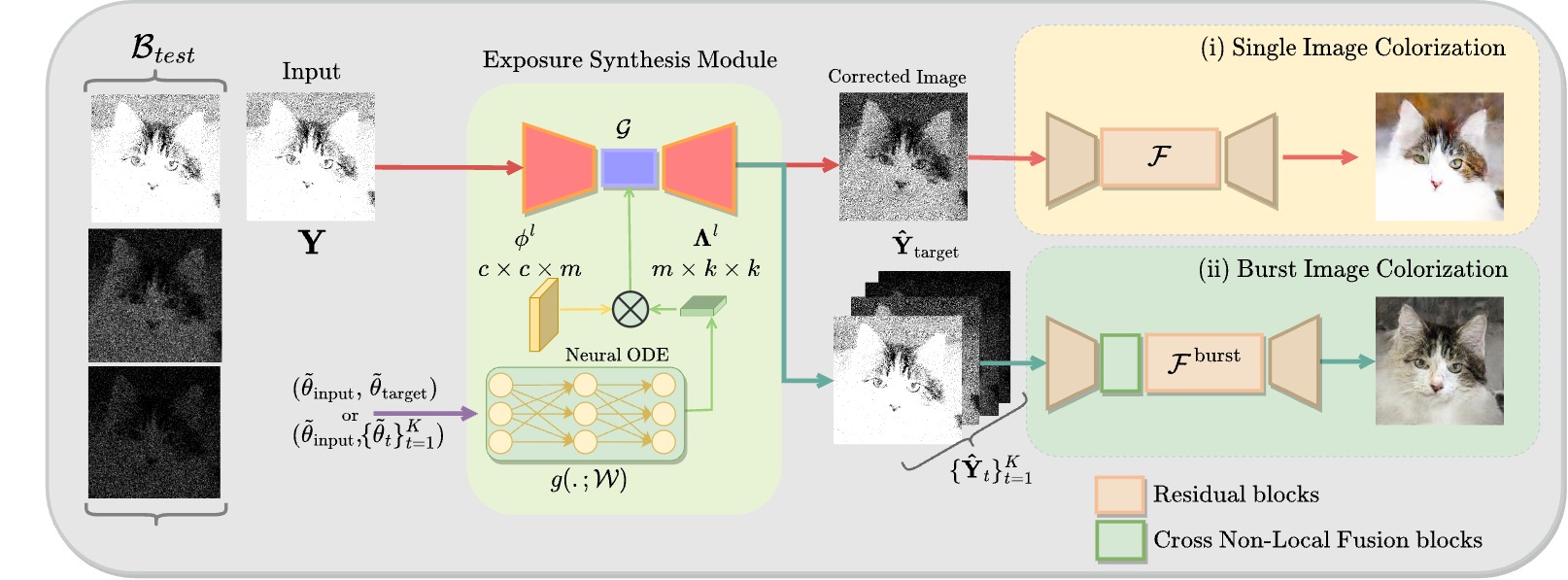}
        \caption{An illustration of our proposed method for exposure adaptive colorization: A binary image, $\vY$, which can be overexposed or underexposed, is input into the proposed exposure synthesis module. The colorization of binary image can be achieved using Single Image Colorization (SIC) or Burst Image Colorization (BIC). (\romannumeral 1) \textbf{SIC:} Based on the input and target exposure levels, $\widetilde\theta_\text{input}$ and $\widetilde\theta_\text{target}$, this module adjusts the weights of a exposure synthesis network $\mathcal{G}$, which then generates an exposure-corrected image. Note that corrected image is not necessarily a binary image. Since the colorization module $\mathcal{F}$ is trained to colorize only image of specific exposure, the exposure synthesis module ensures corrected binary image has similar exposure to the one on which $\mathcal{F}$ is trained. (\romannumeral 2) \textbf{BIC:} For BIC we generate images with varying exposures as input to the burst image colorization network, $\mathcal{F}_{\text{burst}}$. The trained network is able to exploit the complementary information across multiple exposures with the help of Cross Non-Local Fusion blocks~\cite{Luo_2021_CVPR} to synthesize colors in regions of the image that is otherwise not possible by SIC approach.}
        \label{fig:our_method}
    \end{figure*}
    \subsection{Continuous Exposure Synthesis using Filter Decomposed Neural ODEs} 
    \label{subsec:cont_func}
    
    The image formation model defined in Section \ref{subsec:img_formation} indicates that we can formulate the continuum burst of binary images as $\vY(\widetilde{\theta})$, where the binary image $\vY$ is a function of a continuous exposure value $\widetilde{\theta} \in (0, 1]$. A small $\widetilde{\theta}$ value indicates a higher exposed image $\vY$. One of the core contributions of this work is a deep neural network architecture that synthesizes a binary image $\vY_t$ at a target exposure $\widetilde{\theta}_{\text{target}}$ given an input binary image $\vY_i$ measured at the exposure value $\widetilde{\theta}_{\text{input}}$. 

    As the exposure values are continuous, we may need an exhaustive training set that densely samples every possible pair of input exposure value $\widetilde{\theta}_{\text{input}}$ and target exposure value $\widetilde{\theta}_{\text{target}}$ for every scene to train our network. To tackle this problem, we designed a convolutional architecture that enforces the parameter of the network to smoothly change according to the exposures. 

     We use the convolution filter decomposition technique \cite{qiu18dcfnet, wang18-cgan, wang20stochastic} to represent the convolutional filters $\mathbf{F}$ with a low number of parameters, and we leverage neural ODEs~\cite{chen18-node} to associate parameters with image exposures. With these techniques, we enable the adaptive control of filter parameters $\mathbf{F}$ and thereby the learning of the exposure synthesis operation $\mathcal{G}$ efficiently.

     \textbf{Efficient Modelling of Filter Subspace.} Specifically, we formulate the filter parameters $\mathbf{F}$ using the atom-coefficient \cite{qiu18dcfnet} decomposition, which permits the smallest computational bottleneck in modeling filter subspaces among the candidates. Given a convolutional filter $\mathbf{F} \in \mathbb{R}^{c' \times c \times k \times k}$ where $c$, $c'$ and $k$ are the output, input channels, and spatial size, respectively. The atom-coefficient decomposition over a set of $m$ filter atoms decomposes $\mathbf{F}$ into coefficients, denoted by $\mathbf{\phi} \in \mathbb{R}^{c' \times c \times m}$, and atoms, denoted by $\mathbf{\Lambda}\in \mathbb{R}^{m \times k \times k}$
    \begin{align*}
        \mathbf{F} =  \mathbf{\phi}\mathbf{\Lambda}. 
    \end{align*}
   
    When $m \ll c$, the dimension of the parameters is significantly reduced. Extending the decomposition to a L-layer convolutional network gives a set of $L$ filter atoms $\mathbf{\Lambda} = \{\mathbf{\Lambda}^l\}_{l=1}^{L}$ and $L$ coefficients $\phi = \{\phi^l\}_{l=1}^{L}$. 

    \textbf{Enforcing Filter Atom Continuity.} We enforce the filter atoms $\mathbf{\Lambda}$ to smoothly change according to the input and target exposure levels $\widetilde{\theta}_{\text{input}}$ and $\widetilde{\theta}_{\text{target}}$. To do so, we incorporate the continuous-time networks like neural ODEs to describe the relationship of $\mathbf{\Lambda}(\widetilde{\theta}_{\text{input}}, \widetilde{\theta}_{\text{target}})$, as the neural ODEs provide advantages in the number of parameters required, memory consumption, and computational adaptivity~\cite{chen18-node}. The derivative of the filter atoms $d\mathbf{\Lambda}/d\widetilde{\theta}$ w.r.t. the exposure value $\widetilde{\theta}$ as a neural network $g(\cdot)$
    \begin{align*}
        d\mathbf{\Lambda}/d\widetilde{\theta} = g(\widetilde{\theta}_{\text{input}}, \widetilde{\theta}_{\text{target}},\mathbf{\Lambda_{\text{init}}}; \mathcal{W}),
    \end{align*}
    where $\mathbf{\Lambda}_\text{init}$ is the initial state of filter atoms, and $\mathcal{W}$ is the learnable parameters of $g(\cdot)$. The initial condition $\mathbf{\Lambda}_{\text{init}}$ is a learnable parameter and is optimized jointly with $\mathcal{W}$.  Given the input and target exposure values $\widetilde{\theta}_{\text{input}}$ and $\widetilde{\theta}_{\text{target}}$, the corresponding filter atoms can be calculated as
    \begin{equation*}
        \mathbf{\Lambda}(\widetilde{\theta}_{\text{target}}, \widetilde{\theta}_{\text{input}}) =  \mathbf{\Lambda}_\text{init} + \int_{\widetilde{\theta}_{\text{input}}}^{\widetilde{\theta}_{\text{target}}}g(\mathbf{\Lambda}_{\text{init}}, \mathcal{W}) \: d\widetilde{\theta}.
    \end{equation*}
    According to Picard-Lindel\"{o}f existence theorem \cite{coddington}, for a given initial condition, the solution to the IVP is unique. Theoretically, we prove continuity in filter atoms induces controlled changes in image exposure. 

    \begin{theorem} 
    If $\mathbf{\Lambda}_1$ and $\mathbf{\Lambda}_2$ are filter atoms generated with neural ODE using integration intervals ($\widetilde{\theta}_{\text{0}}$, $\widetilde{\theta}_{\text{1}}$) and ($\widetilde{\theta}_0,\widetilde{\theta}_2$), respectively such that $\|\mathbf{\Lambda}_1 - \mathbf{\Lambda}_2  \| \leq \epsilon|\widetilde{\theta}_1 - \widetilde{\theta}_2|$, for $\epsilon > 0$. If $\theta_1$ and $\theta_2$ are the exposures at $\widetilde{\theta}_1$ and $\widetilde{\theta}_2$, and activation function $\sigma$ is non-expansive, then change in exposure $\Delta \theta$ is continuous in filter atoms.
    \label{thm_1}
    \end{theorem}
        
    Details of Theorem \ref{thm_1} and its proof can be found in the Appendix section A1. An illustration of our parameter efficient approach is shown in Figure \ref{fig:our_method}.
    
    \subsection{Exposure Adaptive Colorization} \label{subsec:exp_spectrum}
    Our proposed exposure adaptive colorization consists of an exposure correction realized by the stated synthesis and exposure level specific colorizer.
    \begin{figure}
        \centering
        \includegraphics[width=\columnwidth]{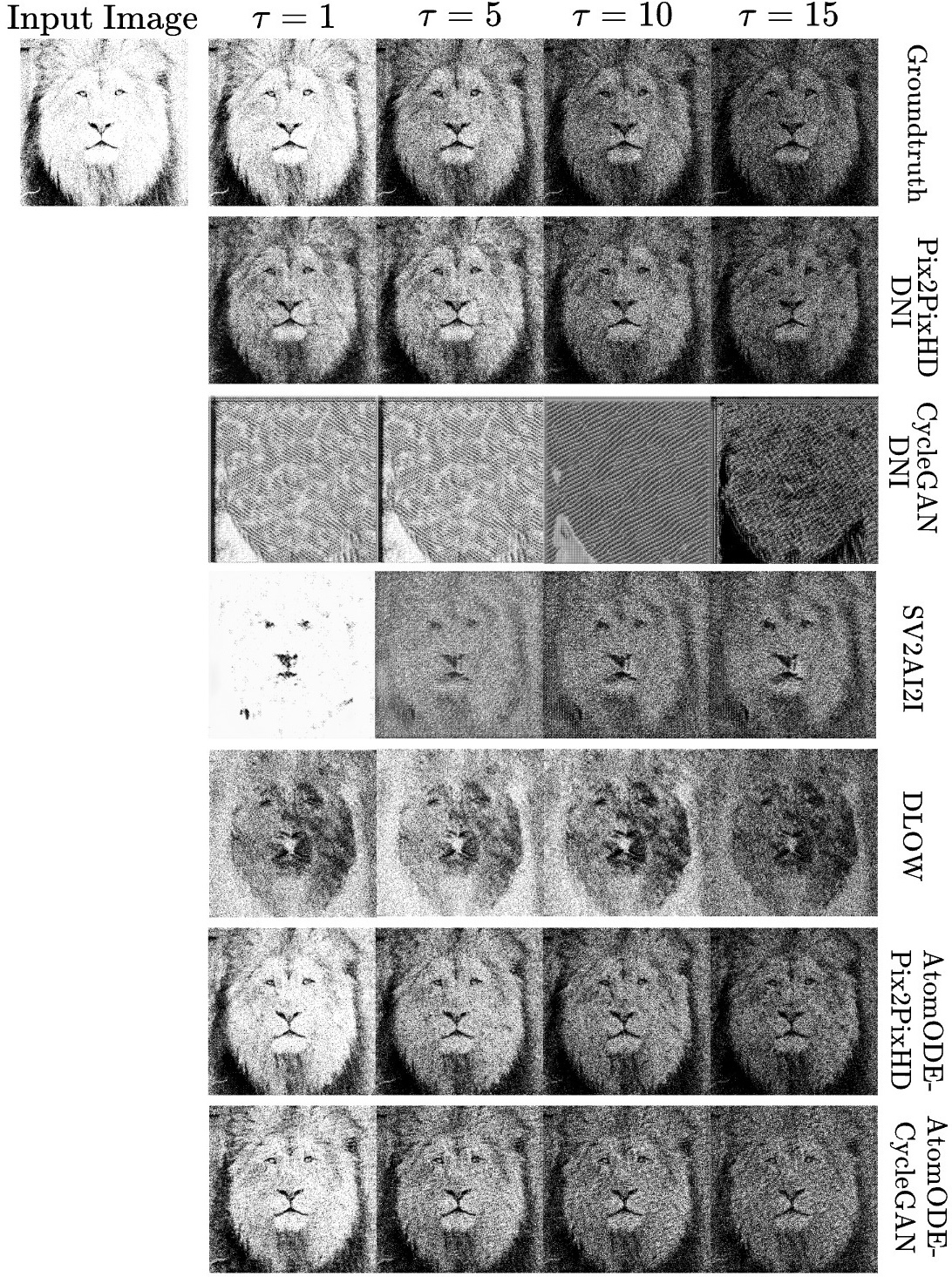}
        \caption{Qualitative results for exposure burst recovery experiments for AFHQ dataset. We compare our methods, AtomODE-Pix2Pix and AtomODE-CycleGAN, with DLOW, SAVI2I and DNI.}
        \label{fig:exposure_burst}
        \vspace{-8mm}
    \end{figure}
    
    \textbf{Exposure correction.} Given a colorization method specific to exposure $\widetilde{\theta}_{\text{target}}$ and a input sample $(\vY_i, \widetilde{\theta}_{\text{input}})$, the proposed exposure correction generates image $\hat{\vY}_t$ at the target exposure in two steps. With the operation $\mathcal{G}$ carried out by some generative network, first, the filter atoms of the network's convolutional filters are computed using the neural ODEs based on the $(\widetilde{\theta}_{\text{input}}, \widetilde{\theta}_{\text{target}})$ pair; next, the convolutional filter weights of the network are updated using the filter atoms and their corresponding coefficients, and the network  
    generate the exposure corrected image $\hat{\vY}_t$ based on $\vY_i$. In the case of burst colorization, multiple images are generated corresponding to a set of target exposures.

     \textbf{Colorization.} Both single-image-based colorization and burst-based colorization are considered in our work. The first approach involves a colorization network $\mathcal{F}: \hat{\vY}_t \rightarrow \mathbf{X}_c$ which takes a exposure corrected binary image as input and outputs corresponding RGB image $\mathbf{X}_c$. The network is trained exclusively at a fixed exposure level, and variations in the exposure of the input image is pre-adjusted by the exposure correction step. The second approach uses a colorization network $\mathcal{F}_{\text{burst}}: \{\hat{\vY}_t\}_{t=1}^{K} \rightarrow \mathbf{X}_c$ that takes in a burst of $K$ binary images $\{\hat{\vY}_t\}_{t=1}^{K}$ corresponding to pre-defined varying exposure values  $\{\widetilde{\theta}_{t}\}_{t=1}^{K}$ and outputs one RGB image $\mathbf{X}_c$.

\begin{table*}[]
\setlength\doublerulesep{1pt}

\resizebox{\textwidth}{!}{%
\begin{tabular}{@{}c|ccc|cc@{}}
\toprule[1pt] \bottomrule[0.3pt]
  \multirow{3}{*}{Exposure Synthesis Method}
 &
  \multicolumn{3}{c|}{AFHQ (512 x 512)} &
  \multicolumn{2}{c}{CelebA-HQ (256 x 256)} \\ \cline{2-4}  \cline{5-6} 
 &
  \multicolumn{1}{c}{\begin{tabular}[c]{@{}c@{}}Cat\\ MSE ($\downarrow$) / RL ($\uparrow$) / FID ($\downarrow$)\end{tabular}} &
  \multicolumn{1}{c}{\begin{tabular}[c]{@{}c@{}}Dog\\ MSE ($\downarrow$) / RL ($\uparrow$) / FID ($\downarrow$)\end{tabular}} &
  \multicolumn{1}{c|}{\begin{tabular}[c]{@{}c@{}}Wild\\ MSE ($\downarrow$) / RL ($\uparrow$) / FID ($\downarrow$)\end{tabular}} &
  \multicolumn{1}{c}{\begin{tabular}[c]{@{}c@{}}Male\\ MSE ($\downarrow$) / RL ($\uparrow$) / FID ($\downarrow$)\end{tabular}} &
  \multicolumn{1}{c}{\begin{tabular}[c]{@{}c@{}}Female\\ MSE ($\downarrow$) / RL ($\uparrow$) / FID ($\downarrow$)\end{tabular}} \\ \hline
Pix2Pix-DNI &
  3.75 / 0.9816 / 61.89 &
  5.91 / 0.9844 / 95.43 &
  8.13 / 0.9780 / 39.29 &
  3.74 / 0.9563 / 56.13 &
  4.78 / 0.9681 / \textbf{39.81} \\
CycleGAN-DNI &
  9.61 / 0.9816 / 207.82 &
  16.79 / 0.9824 / 220.90 &
  9.62 / 0.9820 / 336.38 &
  9.31 / 0.9824 / 94.78 &
  14.26 / 0.9810 / 250.61 \\
SAVI2I &
  41.15 / 46.27 / 178.29 &
  46.27 / 0.9804 / 217.98 &
  38.87 / 0.9019 / 139.47 &
  67.26 / 0.9563 / 111.05 &
  69.21 / 0.9698 / 261.45 \\
DLOW &
  2.44 / 0.9654 / 92.30 &
  2.32 / 0.9489 / 173.65 &
  2.75 / 0.9632 / 179.23 &
  1.87 / 0.9655 / 62.57 &
  1.37 / 0.9798 / 63.31 \\ \hline
\textbf{AtomODE-Pix2Pix (ours)} &
  2.29 / \textbf{0.9997} / 57.88 &
  3.17 /\textbf{ 0.9994} / 140.23 &
  2.15 / \textbf{0.9998 }/ 79.76 &
  2.15 / \textbf{0.9995} / \textbf{53.44} &
  3.20 /\textbf{ 0.9995 }/ 40.08 \\
\textbf{AtomODE-CycleGAN (ours)} &
  \textbf{1.15} / 0.9948 / \textbf{60.54} &
  \textbf{1.18} / 0.9971 / \textbf{87.43} &
  \textbf{1.23} / 0.9979 / \textbf{35.75} &
  \textbf{1.58} / 0.9986 / 77.82 &
  \textbf{1.87} / 0.9994 / 62.59 \\ 
\toprule[0.3pt] \bottomrule[1pt]
\end{tabular}%
}
\vspace{-1ex}
\caption{Colorization results on AFHQ and CelebA-HQ datasets. The input image for all these experiments is an overexposed image at index $\tau=0$ of exposure burst. All the colorizers are trained using an image corresponding to $\tau=10$ in the exposure burst. At test time we correct the overexposed input to correspond to the one used during training. In the first column of the table, we note the combination of the exposure correction method used and colorizer is same across all the methods.}
\vspace{-2.5ex}
\label{tab:expo_burst}
\end{table*}
\section{Experiments}\label{sec:exps}
    \subsection{Experimental Setup}\label{subsec:exp_settings}
    We evaluate the performance of both the exposure synthesis module standalone and the entire proposed approach. We conduct exposure burst recovery experiment to evaluate the exposure synthesis and Single Image Colorization (SIC) and Burst Image Colorization (BIC) experiments to evaluate the final colorization results.

    \textbf{Implementation of Exposure Adaptive Colorization}. We implement the neural ODEs using a 6-layer multilayer perceptron. The generative networks we use for exposure synthesis are variants of Pix2Pix~\cite{pix2pix2017} and CycleGAN~\cite{zhu17}, with which the models are named AtomODE-Pix2Pix and AtomODE-CycleGAN. For colorization, we train a Pix2Pix colorizer dedicated to SIC task and another to BIC. For BIC, the colorizer is additionally equipped with a Cross Non-Local Fusion (CNLF) \cite{Luo_2021_CVPR} to constructively fuse the representations obtained from the burst.
    Note that the colorizers are trained directly on pairs of binary and color image data at fixed $\widetilde\theta$, independent of exposure synthesis. In our colorization experiment, the same trained colorizer is applied on all exposure correction methods for comparison. Details can be found in the Appendix section A2.

    \textbf{Datasets and Synthetic Data Generation.} 
    We train and evaluate the proposed method as well as the baselines on synthetic data generated using images from two datasets: AFHQ \cite{choi2020starganv2} and CelebA-HQ \cite{karras2018progressive}. AFHQ dataset contains images of three classes, cats, dogs, and wild animals, and CelebA-HQ contains images of male and female human faces. 
    
    To synthesize binary images, we use the image formation model described in Section \ref{subsec:img_formation} and apply \textit{exposure bracketing} to generate various exposure. Details of data synthesis can be found in the Appendix section A2. Our compiled test datasets consist of 15-frame bursts of exposure varying from overexposed to underexposed. We index the exposure of these images in a burst by $\tau$, a discretized label of $\widetilde\theta$.
    
\begin{figure*}
      \centering
       \includegraphics[width=\textwidth]{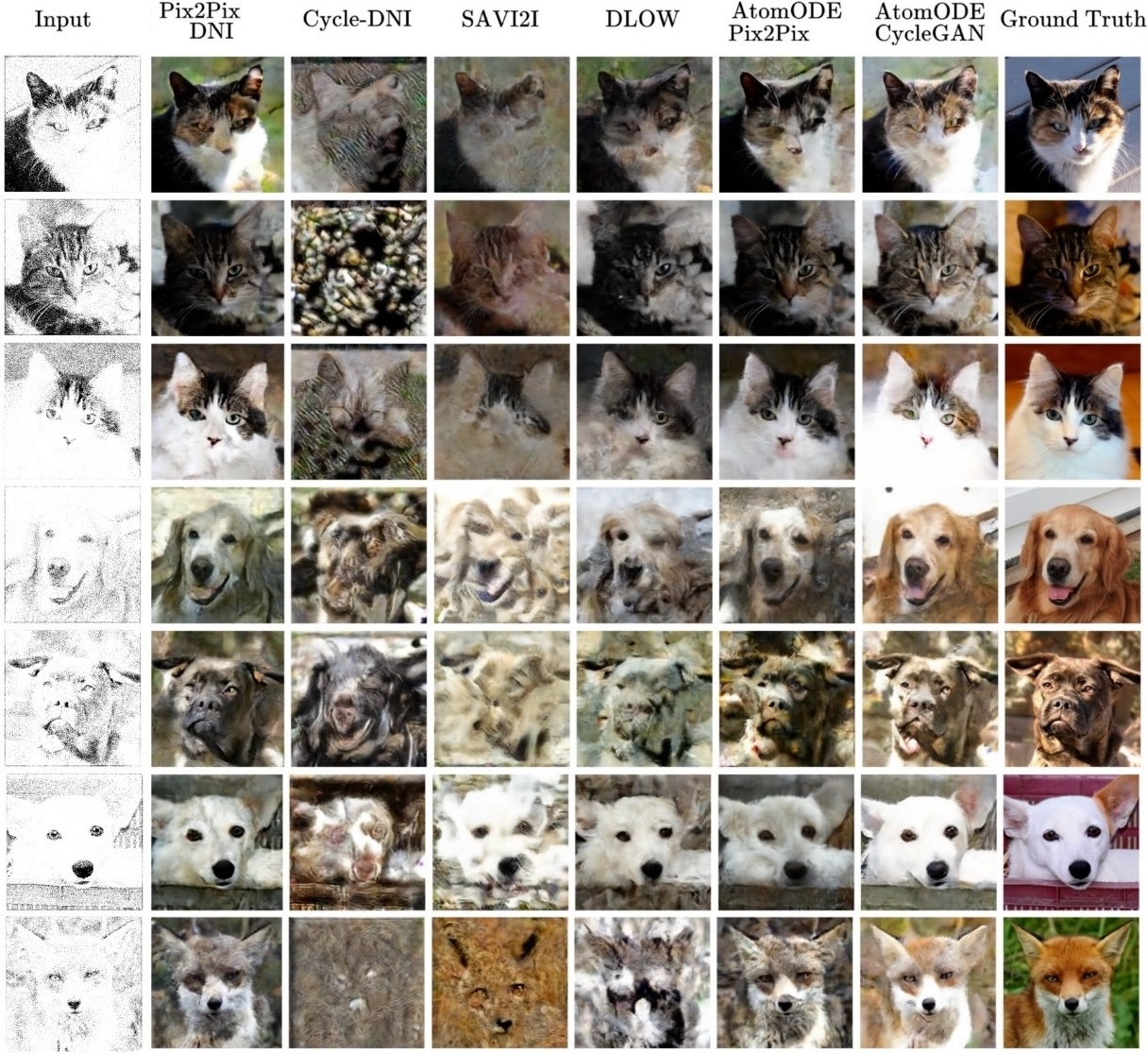}
       \caption{Colorized results for AFHQ dataset using various combinations of exposure correction methods and colorizer. The first column shows the overexposed binary input image for exposure correction. The results of our methods \textbf{AtomODE Pix2Pix} and \textbf{AtomODE CycleGAN} are shown in columns six and seven. }
       \label{fig:colorized_results}
       \vspace{-5mm}
\end{figure*}

    \textbf{Baselines.} In the exposure burst recovery experiment, we compare our approach against DLOW \cite{Gong_2019_CVPR} and SAVI2I \cite{mao2020continuous}. We further consider DNI \cite{wang19} for its capability of \emph{interpolating} network parameters to generate continuous variations of outputs. We prepare baselines Pix2Pix-DNI and CycleGAN-DNI by coupling DNI with Pix2Pix and CycleGAN exposure generators.
    In colorization experiments, we apply the independently trained Pix2Pix colorizers to the exposure correction results generated by all methods.  
    For more details on baseline training and hyperparameters, we refer readers to the Appendix section A2. 
    
    \textbf{Metrics.} In the burst recovery experiment, we choose to use Mean Squared Error (MSE) and Relative Linearity (RL) \cite{lu2022contrastive} to distinguish a smoothly varying exposure burst with consistent appearance changes. A high RL score indicates a better transition in exposure among consecutive samples. In the colorization experiments, we assess the quality of exposure correction by computing the Fr\'echet Inception Distance (FID) between the colorized image and ground truth.
\begin{figure}[t]
      \centering
       \includegraphics[width=\columnwidth]{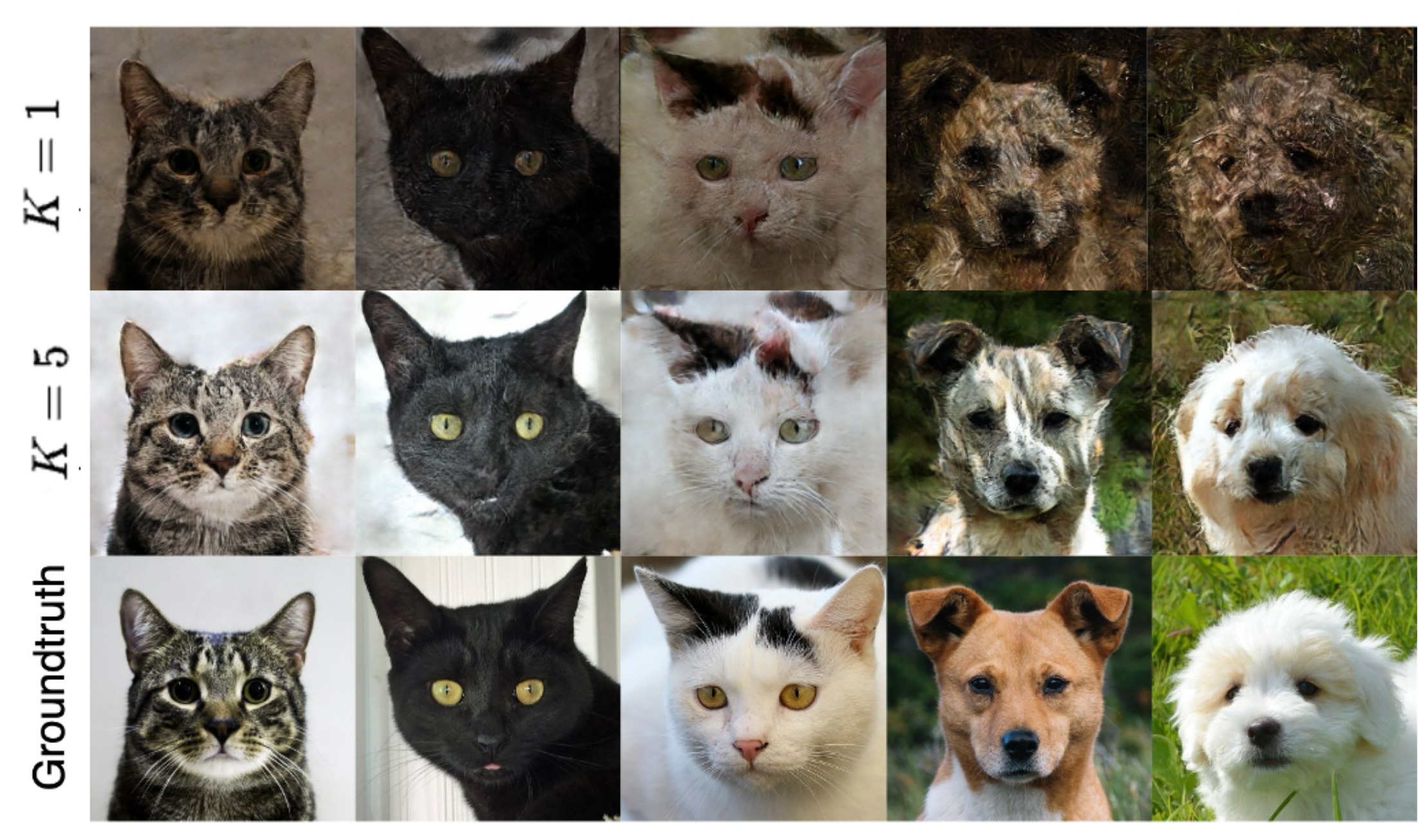}
       \caption{An illustration of burst image colorization for AFHQ dataset using multiple binary images of varying exposures as input. The colorizer is trained using images of exposure $\tau=\{0, 4, 8, 10, 14\}$ . An overexposed input image is used to synthesize multiple images of varying exposures and passed to the colorizer.}
       \label{fig:burst_colorization_results}
\end{figure}
    \subsection{Exposure Burst Recovery Results} \label{subsec:burst_recovery}
    Figure \ref{fig:exposure_burst} presents the qualitative comparison of all methods on AFHQ dataset. We find our methods, AtomODE Pix2Pix and AtomODE CycleGAN are able to generate consistent set of smoothly varying exposure images than baseline methods and match ground truth images generated from exposure bracketing. We also observe some baselines struggle with generation of overexposed frames. All methods for comparison are further assessed on both datasets using MSE and RL, with the quantitative results shown in Table \ref{tab:expo_burst}. Results suggest that the proposed AtomODE provides consistently better performance in all testing conditions across multiple datasets. Additional qualitative results can be found in Appendix section A4.
    \subsection{Single and Burst Image Colorization Results}
\begin{figure}[t]
          \centering
           \includegraphics[width=\columnwidth]{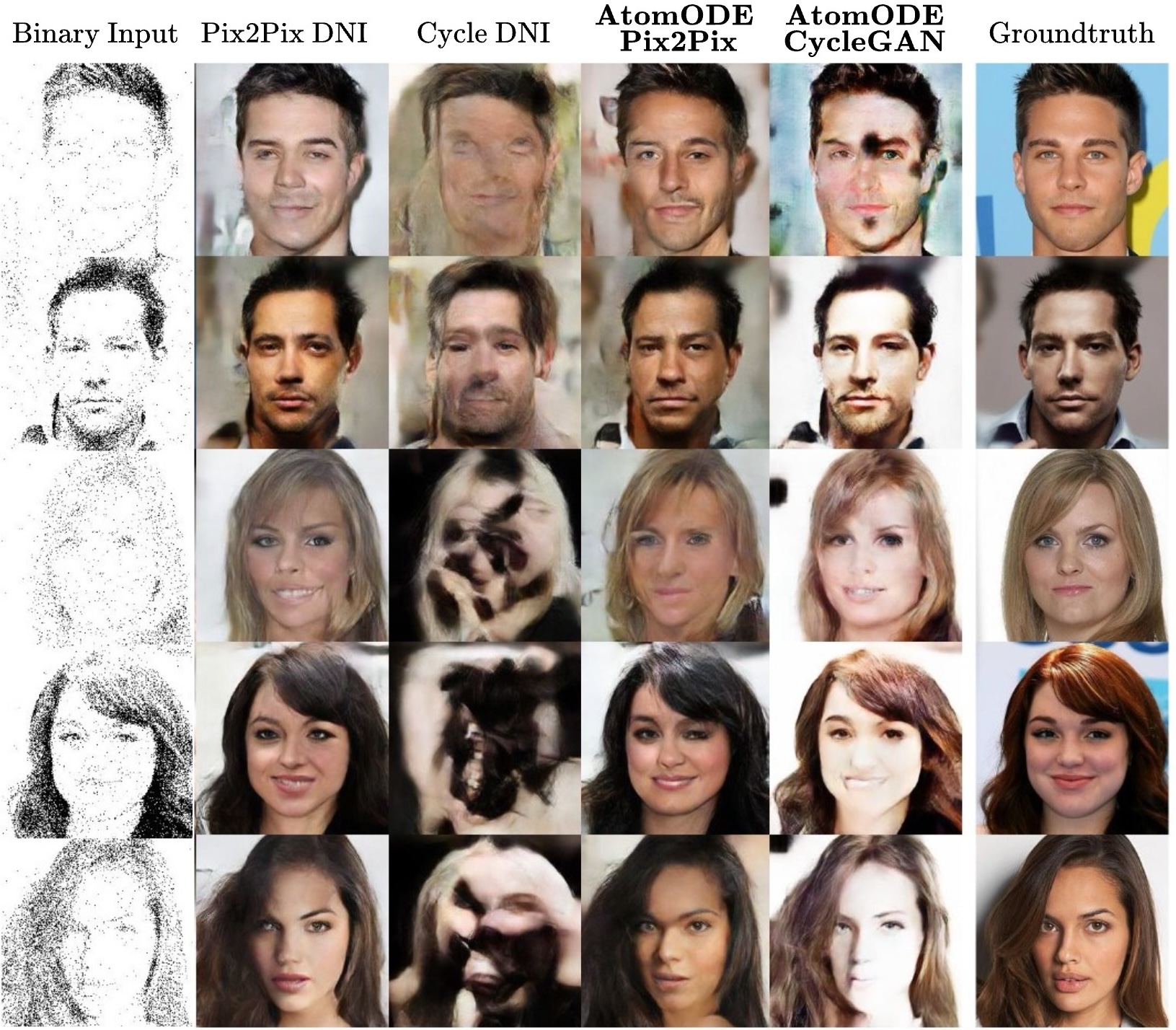}
           \caption{Colorized results for CelebA-HQ dataset using various combinations of exposure correction methods and colorizers. The results of our method are shown in columns four and five.}
           \label{fig:bic_fused}
           \vspace{-3mm}
    \end{figure}

    The SIC colorization results of all methods on AFHQ dataset are shown in Figure \ref{fig:colorized_results}, where the exposure of the test samples is different from the that of the colorizer's training samples. Corrupted colorized images are generated by most of the baselines, while the proposed method produces visually appealing outputs. Figure \ref{fig:bic_fused} shows the SIC colorization results on CelebA-HQ dataset. Additionally, we also present the qualitative results for BIC in Figure \ref{fig:burst_colorization_results} where $K$ denotes the number of binary frames input to the colorizer. The quantitative results are shown in Table \ref{tab:expo_burst}, assessed with the FID metric.

\begin{figure}
      \centering
       \includegraphics[width=\columnwidth]{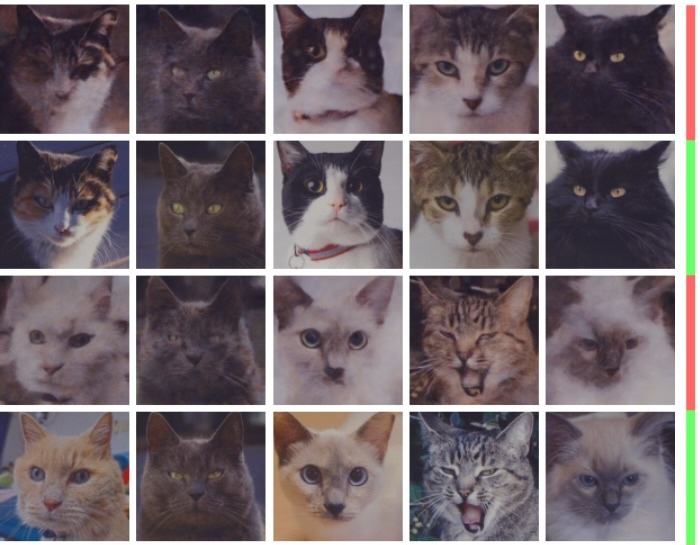}
       \caption{Results on real-world images captured from a CMOS camera (images shown along the row with a red line are colorized images, and along the green line are the ground-truth images captured using an RGB CMOS camera).}
       \label{fig:cmos_results}
\end{figure}
\begin{figure}
      \centering
       \includegraphics[width=\columnwidth]{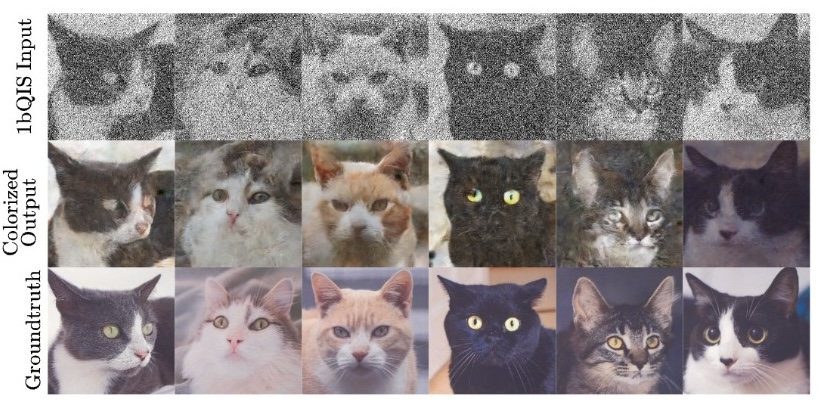}
       \caption{Colorized output using input from prototype QIS camera.}
       \vspace{-3mm}
       \label{fig:qis_results}
\end{figure}
    \subsection{Real Data Experiments} \label{subsec:real_world} 
    As our approach is trained on synthetic data, it is essential to evaluate its performance on real-world data. To this end, we conduct experiments using both a CMOS camera and a prototype QIS camera. We refer the readers to Appendix for further details about the experiment setup. We print 100 images from the training set and 25 from the test set, and we capture pictures of them using both cameras, collecting 125 pairs of grayscale and RGB images. We fine-tune the colorization model with the 100 real data samples, and we evaluate our proposed approach on the set of 25 test data, including both binarized CMOS data and QIS data. In Figure \ref{fig:cmos_results}, we present qualitative results from the evaluation on the real data obtained by the CMOS camera. Despite that the proposed method was not trained with vast real sensor data, it is able to produce visually pleasing colorized images. We observe that fine-tuning the model using a small training dataset significantly improves the colorization quality. We further present results from the evaluation on images captured by the QIS camera in Figure \ref{fig:qis_results}. These results demonstrate the potential of the proposed method for real-world single-photon imaging applications.

\subsection{Benefits for Practical Applications} 
Our methodology is specifically designed with modularity and parameter efficiency in mind, features crucial for real-world industry applications. The Pix2PixDNI method, the second-best performing approach, necessitates training multiple networks and either interpolating model weights on-the-fly during inference or precomputing and storing them to generate images with varied exposures. Conversely, our method circumvents the additional storage requirements and the need to train multiple model parameters by incorporating the exposure correction into the neural ODE framework. 


\section{Conclusion}\label{sec:conclusion}
In this work, we proposed a generative exposure correction approach that can synthesis a spectrum of exposure given a single 1-bit image. We modeled the binary signal variation with respect to exposure change as a continuous function by adapting a small set of parameters of a network, known as filter atoms, to desired exposure using neural ODE. Theoretically and empirically, we demonstrated successful recovery of bursts of 1-bit images from a single input image, and we verified its application on single-image-based and burst-image-based 1-bit image colorization. Experiments on real CMOS and QIS data further indicated the potentials of the proposed approach on real-world single-photon imaging applications.
\clearpage
{
    \small
    \bibliographystyle{ieeenat_fullname}
    \bibliography{main}
}
\end{document}